\documentclass[
twocolumn,
]{ceurart}

\usepackage{hyperref}

\begin{document}

\copyrightyear{2023}
\copyrightclause{Copyright for this paper by its authors.
  Use permitted under Creative Commons License Attribution 4.0
  International (CC BY 4.0).}

\conference{ICAPS 2023: Scheduling \& Planning Applications woRKshop}

\title{Modelling the Spread of COVID-19 in Indoor Spaces using Automated Probabilistic Planning}

\author[]{Mohamed Harmanani}
\address[]{School of Computing, Queen's University, Kingston, Canada}

\begin{abstract}
  The coronavirus disease 2019 (COVID-19) pandemic has been ongoing for around 3 years, and has infected over 750 million people and caused over 6 million deaths worldwide at the time of writing. Throughout the pandemic, several strategies for controlling the spread of the disease have been debated by healthcare professionals, government authorities, and international bodies. To anticipate the potential impact of the disease, and to simulate the effectiveness of different mitigation strategies, a robust model of disease spread is needed. In this work, we explore a novel approach based on probabilistic planning and dynamic graph analysis to model the spread of COVID-19 in indoor spaces. We endow the planner with means to control the spread of the disease through non-pharmaceutical interventions (NPIs) such as mandating masks and vaccines, and we compare the impact of crowds and capacity limits on the spread of COVID-19 in these settings. We demonstrate that the use of probabilistic planning is effective in predicting the amount of infections that are likely to occur in shared spaces, and that automated planners have the potential to design competent interventions to limit the spread of the disease. Our code is fully open-source and is available at: \href{https://github.com/mharmanani/prob-planning-covid19}{https://github.com/mharmanani/prob-planning-covid19}.
\end{abstract}

\maketitle

\section{Introduction}
\label{sec:introduction}
Epidemic modelling is a crucial element of the response to any disease's spread, as it may help to estimate the number of infected persons, the number of deaths, and the potential damage the disease may inflict on the economy and the healthcare system. Thus, it is imperative to design models that are as reliable as possible in estimating the spread of the disease, and malleable enough to adapt to new circumstances, given the rapidly evolving nature of pandemics. 

The applications of automated planning to the field of epidemic modelling have been studied, but remain under-explored nonetheless. Moreover, most of the work involving COVID-19 modelling has focused on stopping or minimizing the spread of the disease in a large population such as cities \cite{ThomazMauáBarros2020, LeungDingHuangRabbany2020}, states and provinces \cite{HousniSumidaRusmevichientongTopalogluZiya2020}, countries \cite{OgdenFazilArinoBerthiaumeFismanGreerLudwigNgTuiteTurgeonetal2020}, or even the world. The purpose of this work is to show the effectiveness of automated probabilistic planning on a smaller scale, like modelling transmission in shared spaces like households, workplaces, or schools. We show that indoor interactions can be modelled on a grid-based structure using RDDL and the PROST planner, and that probabilistic planning can be applied to those grids to predict the potential of infection of individuals sharing a particular space. 

\section{Background}
\label{sec:background}
\begin{figure}[htbp!]
    \centering
    \includegraphics[width=0.75\columnwidth]{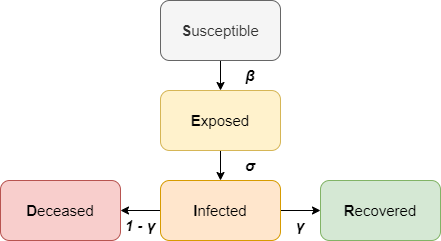}
    \caption{State transitions for any given individual as described by SEIRD}
    \label{fig:seird}
\end{figure}

\paragraph{SEIR Modelling} The SEIRD model is an extension of the traditional SEIR model \cite{aronSEIR}, a general technique used to model the progression of a disease in a society. In this model, individuals are grouped into five categories:
\begin{itemize}
    \item \textbf{S}usceptible individuals at any given time $t$, which have not yet been exposed or infected
    \item \textbf{E}xposed individuals, which have interacted with an infected person, but do not show symptoms yet
    \item \textbf{I}nfected individuals, who have the disease and have shown symptoms
    \item \textbf{R}ecovered individuals, who have had the disease but recovered from it, and no longer show any symptoms. In the traditional model, these individuals cannot get reinfected.
    \item \textbf{D}eceased individuals, who did not recover from the disease
\end{itemize}
Figure \ref{fig:seird} and the following equations describe the dynamics of the model at each timestep:
\begin{align*}
    \frac{dS}{dt} &= -\beta\frac{S}{N}I\\
    \frac{dE}{dt} &= \beta\frac{S}{N} - \sigma E \\
    \frac{dI}{dt} &= \sigma E - \gamma I\\
    \frac{dR}{dt} &= \gamma I\\
    \frac{dD}{dt} &= (1-\gamma)I
\end{align*}
where $\beta$, $\sigma$, and $\gamma$ are the transition rates from S to E, E to I, and I to R respectively. The transition rate from I to D is given by $(1-\gamma)$. 

 \section{Related Work}
\label{sec:related_work}

\paragraph{Contact networks}
There have been various works that leverage the structure of contact networks to estimate the spread of COVID-19. Aparicio and Pascual use small world networks with a classical SIR \cite{KermackMcKendrickWalker1997} epidemic model \cite{AparicioPascual2006}. Hoen et al. use public Wifi networks to estimate physical contact networks, and identify robust community structures in which epidemics are likely to spread \cite{HoenHladishEggoLencznerBrownsteinMeyers2015}. More recently, Leung et al. introduced the Contact Graph Epidemic Model (CGEM) for SEIR modelling. They use the Montreal Wifi network's connection records to generate their contact network, and obtain realistic results that are more in-line with the numbers provided by public health authorities when compared to that of other similar models \cite{LeungDingHuangRabbany2020}.

\paragraph{Automated planning} 
There are several works that make use of planning techniques to model disease spread. Thomaz et al. \cite{ThomazMauáBarros2020} use MDPs in conjunction with a contact network structure of the city of S\~{a}o Paulo to model the spread of the disease. They also investigate a multitude of NPIs, as well as different levels of lockdown intensity, and their impact on the severity of the epidemic. Xue models the problem of conservation planning as a diffusion network, and utilizes automated planning techniques such as Hindsight Optimization (HOP) as a solution \cite{Xue2020}. This approach is then evaluated on epidemic modelling problems, with promising results \cite{Xue2020}.  
Kinathil et al. utilize Parametrized Hybrid MDPs (PHMDP) to solve a variety of problems. One of the domains they apply their model to is that of SIR epidemic modelling \cite{kinathil2017nonlinear}. They show that this approach can solve any SIR model without needing an analytical solution. Arruda et al. use parsimonious modelling in conjunction with MDPs to establish a trade-off between the economic damage caused by the virus and the strain imposed on the healthcare system by the number of infected persons \cite{arruda2022}. Li et al. use MDPs to model the problem of contact tracing and isolating infected persons, then use a greedy approximation of a linear program, with promising results in a variety of realistic scenarios \cite{li2021mdpLP}. There have also been applications of planning applied to the broader context of the COVID-19 pandemic, either for resource management \cite{LeeLeeMogkGoldsteinBibliowiczBrudyTessier2021, MarkhorstZverMalbasicDijkstraOttovanderMeiMoeke2021}, or mobile robotics \cite{HeCaoYe2022,BarnawiChhikaraTekchandaniKumarBoulares2021}. 

\section{Methods}
\label{sec:approach}
\subsection{Environment modelling}
\begin{figure}[htbp!]
    \centering
    \includegraphics[width=0.5\columnwidth, height=0.5\columnwidth]{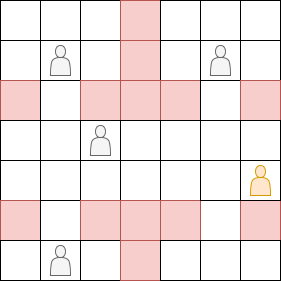}
    \caption{A model of 4 rooms separated by walls (red) and a hallway. The space is populated by 5 people, and 1 person is infected (orange).}
    \label{fig:domain_ex}
\end{figure}

\paragraph{Modelling the space} We use the RDDL programming language \cite{Sanner:RDDL} for our experimental setup. First, we model the space as one large $M \times N$ grid. We represent the grid using $M\cdot N$ location nodes, which are connected to one another by the \texttt{LINK} property, and by an additional directional property (e.g. \texttt{LEFT}, \texttt{RIGHT}, \texttt{DOWN}, \texttt{UP}) indicating the relative position of one tile with regards to another. We divide the space into rooms connected by hallways by making some of the locations inaccessible, using the \texttt{WALL} property. This idea is further illustrated in Figure \ref{fig:domain_ex}. Each person is located on one of the walkable tiles in the grid, modelled by the \texttt{at(?p, ?loc)} fluent. At every timestep, the location of a person may change, as they move from their current location to a new one with probability $p_{mv}$, provided it is accessible, unoccupied by another individual, and connected to their current location. 

\paragraph{Probability of exposure}  The distance between susceptible and infected people will affect the probability of exposure to the virus. For instance, if an infected person moves closer to a susceptible one, the probability of the former infecting the latter increases. If the susceptible individual gets closer to several infected people, that probability increases further. To calculate the exposure probability $\phi$ of a single individual $p$,  we multiply the parameter $\beta$ by $k \in (0, 1]$, representing the average exposure distance needed for the virus to transmit from one person to the other. Hence, we compute $\phi$ as follows
\begin{align*}
    \phi(p) &= \beta \cdot k
\end{align*}

\paragraph{Probability of infection} People exposed to the SARS-CoV-2 virus can take as much as 14 days to begin to show symptoms, but typically take 3 to 7 days to do so. In a traditional SEIRD model, the parameter $\sigma$ describes the rate at which people transition from exposed to infected, and can be computed as the inverse of the time taken for symptoms to manifest, which in this case would be any of $1/3$, $1/7$, or $1/14$. However, because we wish to model the transition from the exposed state to an infected state as a probabilistic event, we find that using these values as probabilities would not accurately reflect the dynamics of COVID-19 transmission. Instead, we set the value of $\sigma$ to 0.95, as we assume that 95\% of exposures lead to a symptomatic infection. After a person gets infected, they have an 80\% probability of staying infected when the next time step comes around. We chose this value based on the length of the self-isolation period mandated by the CDC, which is 5 days \cite{cdc}, i.e. $1 - \frac{1}{5} = 0.8$.

\begin{table}
    \begin{tabular}{c|c|c}
        \hline
         \bf Param. & \bf Value & \bf Description\\
         \hline
         $N$ & varies & Total population\\
         $S_0$ & $N - 1$ & Susceptible persons at $t=0$\\
         $E_0$ & $0$ & Exposed persons at $t=0$\\
         $I_0$ & $1$ & Infected persons at $t=0$\\
         $R_0$ & $0$ & Recovered persons at $t=0$\\
         $D_0$ & $0$ & Deceased persons at $t=0$\\
         $\beta$ & $0.78$ & Transmission rate\\
         $\sigma$ & $0.95$ & Infection probability\\
         $\gamma$ & $0.93$ & Recovery rate\\
         $\mu$ & $0.07$ & Mortality rate\\
         $G_x$ & varies &  Grid dimensions ($x$)\\
         $G_y$ & varies &  Grid dimensions ($y$)\\
         $W_G$ & varies & Number of walkable tiles \\
         $k$ & $1.00$ & Avg. exposure distance \\
         \hline
    \end{tabular}
    \caption{Parameters of the model. The majority of these parameters remain fixed in our experiments, except for $N, G_x, G_y, W_G$.}
    \label{tab:params_table}
\end{table}

\subsection{Planner actions \& goal}
We use the PROST planner to run our simulations. The goal of the planner is to minimize the number of infected and deceased persons. To do so, it may intervene and put constraints in place to curb the spread of the disease. Each one of these interventions will be represented by an action, with a cost and a probability of success. 

\paragraph{Masking} The action \texttt{mask} is used by the planner to enforce a mask mandate on the individuals. Each individual has a probability of non-compliance that we set to 4\%, in accordance with real-world statistics \cite{haischer2020wearing}. To simulate the protective effects of masks, we decrease the probability of transmission when mask mandates are in effect, as done in \cite{LeungDingHuangRabbany2020}. When an individual is wearing a mask, their probability of getting infected is multiplied by 0.8, and the probability of them infecting someone else is multiplied by 0.6. 

\paragraph{Vaccinations} The planner can also \texttt{vaccinate} individuals, as long as they aren't already exposed or infected. As with masks, we introduce a probability of non-compliance of 7\% based on the statistics outlined in \cite{gravelle2022estimating}. For simplicity, we assume that one dose is enough, and that immunity conferred by the vaccine does not wane. However, vaccinated individuals can still transmit the disease, and can get reinfected with a small probability. This serves to counteract the downsides of the assumptions made earlier. Drawing on the work of Lewis et al. \cite{lewis2022effectiveness}, we set the probability of reinfection to
\begin{align*}
    \phi_{vax} &= 0.13 \cdot \phi
\end{align*}
In addition to protecting from infection, vaccines also protect against severe disease and death, so the probability $\mu$ of an individual succumbing to the disease is also multiplied by the protection factor above, improving their chances of survival.

\paragraph{Reward} Let $I_f$, $D_f$ be the sets of all exposed, infected, and deceased individuals respectively at the end of the simulation, and let \texttt{PEN-I}, \texttt{PEN-D} be penalties incurred to the model reward function for each exposed or infected individual respectively. 
The model's reward function can then be expressed as
\begin{align*}
    \texttt{reward} &= I_f\cdot\texttt{PEN-I} + D_f \cdot\texttt{PEN-D}
\end{align*}
The penalty values are all negative. The model's goal is thus minimizing infections and deaths.

\section{Evaluation}
\label{sec:evaluation}
\begin{figure}[htbp!]
    \centering
    \includegraphics[width=0.9\columnwidth]{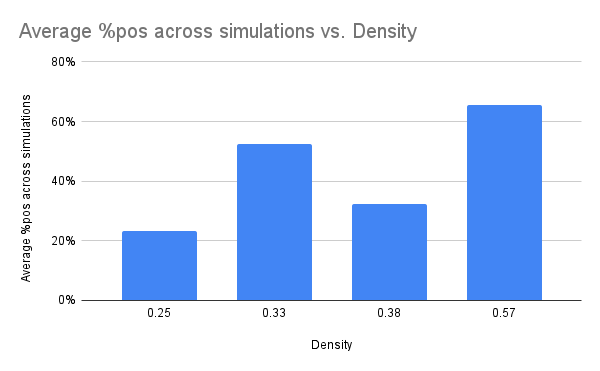}
    \caption{The average percentage of cases plotted against the room density $N/W_G$. The percentage values are obtained by averaging the \%pos. values in Table \ref{tab:results1} across multiple runs and parameter values.}
    \label{fig:compare_rows_1}
\end{figure}

\paragraph{Experimental Setting} We evaluate our planner's predictions when confronted with a variety of factors like the size of the space, the number of people present, and the presence of masks and vaccines. For the room sizes, we use grids of size $4 \times 4$ and $6 \times 6$ to model two rooms of different sizes. We use the quantity $N/W_G$ to measure the density of the room, where $N$ and $W_G$ are the population size and number of walkable tiles respectively. The closer this ratio is to 0, the less crowded the room is.
We also experiment with the presence of masks and vaccines. We run different variations of the simulation. In the first variation, no masks are worn, no one is vaccinated, and the probability of non-compliance for both is set to 1. In other words, the planner can do nothing. In the second variation, masks are worn but no vaccines are available. The planner may use the $\texttt{mask}$ action, but any attempts to $\texttt{vaccinate}$ people will fail. Finally, in the third variation, both masks and vaccines are available, and the planner has access to both actions. The full summary of parameters used as well as the results of our experiments are highlighted in Table \ref{tab:results1}. We find that the planner predicts a small decrease in infections when masks are introduced, and a much steeper one when vaccines are introduced. The average number of deaths also decreases as less people get infected and more people get vaccinated. Furthermore, as evidenced by our results in Table \ref{tab:results1} and Figure \ref{fig:compare_rows_1}, the planner predicts an increase in infections as the room density increases. This prediction is consistent with the observed fact that viruses spread more easily in smaller spaces with more people present.

\begin{table*}[htbp!]
    \centering
    \begin{tabular}{c|ccccccccc}
    \hline
    \textbf{Simulation} & $N$&\textbf{Masks} & \textbf{Vaccines} & \textbf{Walkable Tiles} & $N/W_G$ & \textbf{Pred. \%pos.} & $D_{avg}$\\
    \hline
    Small space & 4 &  No & No & 12/16 & 0.33 & 70\% & 0.90\\
    Small + masks & 4 & Yes & No & 12/16 & 0.33 & 57\%& 0.77 \\
    Small + masks + vaccines & 4 &  Yes & Yes & 12/16 & 0.33 & 30\% & 0.43 \\
    \hline
    Larger space & 8 &  No & No &  32/36 & 0.25 & 33\% & 1.09\\
    Larger + masks & 8 &  Yes & No & 32/36 & 0.25 & 21\% & 0.22 \\
    Larger + masks + vaccines & 8 &  Yes & Yes & 32/36 & 0.25 & 16\% & 0.11 \\
    \hline
    Small and crowded & 8 &  No & No &  14/16 & 0.57 & 85\% & 2.43\\
    Small + crowd + masks + vaccines & 8 &  Yes & Yes & 14/16 & 0.57 & 44\% & 0.90 \\
    \hline
    Larger and crowded & 12 &  No & No &  32/36 & 0.38 & 50\% & 2.00\\
    Larger + crowd + masks + vaccines & 12 &  Yes & Yes & 32/36 & 0.38 & 14\% & 0.00 \\
    \hline
    \hline
    \end{tabular}
    \caption{Summary of the parameters used for each simulation. We run each simulation 3 times, with 5 rounds and 15 timesteps, and one initial infection for each MDP. Walkable tiles are expressed as a fraction of the total grid size, where the remaining tiles are walls. We report the percentage of infections and the average number of deaths predicted by each model. }
    \label{tab:results1}
\end{table*}

\begin{figure}[htbp!]
    \centering
    \includegraphics[width=0.9\columnwidth]{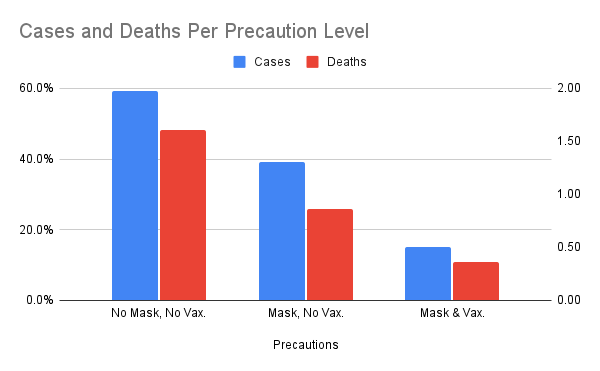}
    \caption{The average percentage of cases and the average number of deaths plotted against the level of precaution taken in each room in Table \ref{tab:results1}. We average the \%pos. and $D_{avg}$ values in the table across multiple runs and precaution levels.}
    \label{fig:compare_rows_2}
\end{figure}

\paragraph{Real-world Benchmarks} We also wish to compare our planner's predictions with real data of COVID-19 outbreaks in a shared indoor setting. To that end, we extract the total number of confirmed COVID-19 cases in specific Ontario public schools between September 2021 and December 2021, as reported by the Government of Ontario \cite{ontarioschools}. While it not possible to know the exact policies employed by each school regarding masks and vaccine, it is worth noting that a mask and vaccine mandate was in effect in Ontario public schools during the chosen period \cite{cihisite, cihi}. A more detailed list of the restrictions imposed during this period can be found in Table \ref{tab:on-interventions-2021}. We extract the enrollment numbers from the Ontario Education Ministry's Open Data Portal \cite{ontarioeducationfinder}, and choose the \'Ecole \'El\'ementaire Catholique du Bon Berger (EECBB) and the AM Cunningham Junior Public School (AMCPS) to serve as baseline comparisons. EECB has 105 total students, with 22 reported infections (21.2\% positivity), and AMCPS has 350 total students, with 58 reported infections (16.6\% positivity).
\iffalse
The main criteria for selecting each school were the total number of students and number of infections. The data extracted is shown in Table \ref{tab:schools}. 
\begin{table}[htbp!]
    \centering
    \begin{tabular}{c|cc}
    \hline
    \textbf{Baseline} & \textbf{Enrollment} & \textbf{Total Infections} \\
    \hline
    EECBB & 105 & 22 \\
    AMCPS & 350 & 58\\
    \hline
    \end{tabular}
    \caption{Summary of the enrollment and total number of infections reported in the schools used as baselines.}
    \label{tab:schools}
\end{table}
\fi
\\

\noindent To simulate a school, we model a single classroom as a $G_x \times G_y$ grid, and place some $m_p$ students in the room as done earlier. For each baseline with enrollment $N_e$, we run the simulation $\lfloor{N_e}/{m_p}\rceil$ times to simulate a school with the appropriate number of students. The number of students modelled by the MDP is thus given by
\begin{align*}
    N_{est} &= m_p \cdot \lfloor{N_e}/{mp}\rceil 
\end{align*}

\noindent For EECB, we set $m_p = 8$ and $G_x = G_y = 6$. For AMCPS, we set $m_p = 17$, $G_x  = 9$ and $G_y = 7$. We then aggregate the number of active cases for each timestep over all runs and measure the relative error when compared to the baseline. We simulate a variety of settings as done in Table \ref{tab:results1}, and report our results in Table \ref{tab:results2}. For both baselines, the models with masks and vaccines achieve the lowest error rates, with the EECB MDP achieving an error rate as low as 0.6\%, and the AMCPS MDP performing reasonably well with an error rate of 9.2\%.  Since we assume that both schools are in compliance with the measures shown in Table \ref{tab:on-interventions-2021}, then we conclude that the planner is capable of accurately predicting the number of infections, with an error rate below 10\% for both baselines.

\begin{table*}[htbp!]
    \centering
    \begin{tabular}{c|ccccc|ccc}
    \hline
    \textbf{Model} & \textbf{Simulations} & \textbf{Masks} & \textbf{Vaccines} & $N$ & $N_{est}$ & \textbf{Pred. \%pos} & \textbf{True \%pos} & \textbf{Abs. Error}\\
    \hline
    EECB-MDP-1 & 13 & No & No & 105 & 104 & 79.8\% & 21.2\% & 59.0\% \\
    EECB-MDP-2 & 13 & Yes & No & 105 & 104 & 53.6\% & 21.2\% & 32.4\% \\
    EECB-MDP-3 & 13 & Yes & Yes & 105 & 104 & 21.8\% & 21.2\% & \textbf{0.6\%} \\
    \hline
    AMCPS-MDP-1 & 21 & No & No & 350 & 357 & 64.1\% & 16.6\% & 47.5\% \\
    AMCPS-MDP-2 & 21 & Yes & No & 350 & 357 & 63.3\% & 16.6\% & 46.7\% \\
    AMCPS-MDP-3 & 21 & Yes & Yes & 350 & 357 & 25.8\% & 16.6\% & \textbf{9.2\%} \\
    \hline
    \hline
    \end{tabular}
    \caption{Evaluating our planner's performance on 3 baseline tasks, with and without masks and vaccines. For each baseline task, we highlight the planner with the lowest error rate. }
    \label{tab:results2}
\end{table*}

\section{Discussion}
\label{sec:Discussion}
\paragraph{Limitations} Because the principal goal of this work is exploring the applications of automated planning to the novel domain of disease transmission on a small scale, we make several assumptions about the dynamics of COVID-19 to simplify the model and planner. For example, we modify some of the SEIRD model's parameters to be more suitable to our probabilistic setting rather than use the true values. It is possible that some of the assumptions made may not be sufficient to capture the full picture of the transmission dynamics of COVID-19, or that may become less accurate with time as the pandemic evolves. Furthermore, we assumed that the public schools used as benchmark tasks were in compliance with government-imposed restrictions such as masks and vaccine mandates. However, there are factors other than compliance that affect the number of cases in a school, such as the fraction of students attending classes remotely. As such, a more comprehensive dataset of COVID-19 indoor transmissions is needed to fully validate the models proposed in this work. 

\begin{table}[ht!]
    \centering
    \begin{tabular}{p{0.18\columnwidth}|p{0.70\columnwidth}}
        \hline
        \textbf{Date} & \textbf{Intervention} \\        
        \hline
        2021-09-02 & Return to in-person learning with an option of synchronous remote learning, maintaining physical distancing and requiring masks\\
        \hline
        2021-09-07 & Schools open for the 2021–2022 school year\\
        \hline
        2021-09-22 & Proof of vaccination in indoor public settings is made mandatory\\
        \hline
        2021-10-05 & Rapid antigen screening is made available for students in public health units where risk of transmission is high\\
        \hline
        2021-12-19 & 50\% capacity limit on indoor public settings (with exceptions) to limit the spread of the Omicron variant\\
        \hline
    \end{tabular}
    \caption{A timeline of the key measures taken by the Government of Ontario to limit the spread of COVID-19 in indoor settings between September and December 2021 \cite{cihisite, cihi}.}
    \label{tab:on-interventions-2021}
\end{table}

\paragraph{Future Work} There are several avenues that may be explored in order to improve the robustness of this approach. First, we may consider using MDPs with a longer horizon and more timesteps to simulate the impact of specific plans and policies over a longer period of time (days, weeks, months, etc...). Moreover, one may consider generalizing the domain further for it to be applicable to a broader range of diseases and settings by adding more actions, state transitions, and parameters. Finally, it may be worthwhile to expand the sizes of the grids used for simulating a space, and compare the performance of this approach on more benchmark datasets.

\section{Conclusion}
\label{sec:conclusion}
We present a novel approach to the modelling of COVID-19 transmission in shared spaces using probabilistic planning by approximating the SEIRD model of disease. We model these spaces as grid structures and use RDDL and the PROST planner to simulate an outbreak and predict the number of infections. Our experiments demonstrate the feasibility of this method, as well as its potential to model the dynamics of indoor transmission during a pandemic.


\begin{thebibliography}{25}
\expandafter\ifx\csname natexlab\endcsname\relax\def\natexlab#1{#1}\fi
\providecommand{\url}[1]{\texttt{#1}}
\providecommand{\href}[2]{#2}
\providecommand{\path}[1]{#1}
\providecommand{\DOIprefix}{doi:}
\providecommand{\ArXivprefix}{arXiv:}
\providecommand{\URLprefix}{URL: }
\providecommand{\Pubmedprefix}{pmid:}
\providecommand{\doi}[1]{\href{http://dx.doi.org/#1}{\path{#1}}}
\providecommand{\Pubmed}[1]{\href{pmid:#1}{\path{#1}}}
\providecommand{\bibinfo}[2]{#2}
\ifx\xfnm\relax \def\xfnm[#1]{\unskip,\space#1}\fi
%Type = Inproceedings
\bibitem[{Thomaz et~al.(2020)Thomaz, Mauá, and Barros}]{ThomazMauáBarros2020}
\bibinfo{author}{G.~Thomaz}, \bibinfo{author}{D.~Mauá},
  \bibinfo{author}{L.~Barros},
\newblock \bibinfo{title}{A contact network-based approach for online planning
  of containment measures for covid-19},
\newblock in: \bibinfo{booktitle}{Anais do Encontro Nacional de Inteligência
  Artificial e Computacional (ENIAC)}, \bibinfo{publisher}{SBC},
  \bibinfo{year}{2020}, p. \bibinfo{pages}{234–245}. \URLprefix
  \url{https://sol.sbc.org.br/index.php/eniac/article/view/12132}.
  \DOIprefix\doi{10.5753/eniac.2020.12132}.
%Type = Article
\bibitem[{Leung et~al.(2020)Leung, Ding, Huang, and
  Rabbany}]{LeungDingHuangRabbany2020}
\bibinfo{author}{A.~Leung}, \bibinfo{author}{X.~Ding},
  \bibinfo{author}{S.~Huang}, \bibinfo{author}{R.~Rabbany},
\newblock \bibinfo{title}{Contact graph epidemic modelling of covid-19 for
  transmission and intervention strategies}  (\bibinfo{year}{2020}). \URLprefix
  \url{http://arxiv.org/abs/2010.03081}.
  \DOIprefix\doi{10.48550/arXiv.2010.03081}, \bibinfo{note}{arXiv:2010.03081
  [physics]}.
%Type = Article
\bibitem[{Housni et~al.(2020)Housni, Sumida, Rusmevichientong, Topaloglu, and
  Ziya}]{HousniSumidaRusmevichientongTopalogluZiya2020}
\bibinfo{author}{O.~E. Housni}, \bibinfo{author}{M.~Sumida},
  \bibinfo{author}{P.~Rusmevichientong}, \bibinfo{author}{H.~Topaloglu},
  \bibinfo{author}{S.~Ziya},
\newblock \bibinfo{title}{Future evolution of covid-19 pandemic in north
  carolina: Can we flatten the curve?}  (\bibinfo{year}{2020}). \URLprefix
  \url{http://arxiv.org/abs/2007.04765}, \bibinfo{note}{arXiv:2007.04765
  [q-bio]}.
%Type = Article
\bibitem[{Ogden et~al.(2020)Ogden, Fazil, Arino, Berthiaume, Fisman, Greer,
  Ludwig, Ng, Tuite, Turgeon, and
  Waddell}]{OgdenFazilArinoBerthiaumeFismanGreerLudwigNgTuiteTurgeonetal2020}
\bibinfo{author}{N.~H. Ogden}, \bibinfo{author}{A.~Fazil},
  \bibinfo{author}{J.~Arino}, \bibinfo{author}{P.~Berthiaume},
  \bibinfo{author}{D.~N. Fisman}, \bibinfo{author}{A.~L. Greer},
  \bibinfo{author}{A.~Ludwig}, \bibinfo{author}{V.~Ng}, \bibinfo{author}{A.~R.
  Tuite}, \bibinfo{author}{P.~Turgeon}, \bibinfo{author}{L.~A. Waddell},
\newblock \bibinfo{title}{Modelling scenarios of the epidemic of covid-19 in
  canada},
\newblock \bibinfo{journal}{Canada Communicable Disease Report}
  (\bibinfo{year}{2020}) \bibinfo{pages}{198–204}.
  \DOIprefix\doi{10.14745/ccdr.v46i06a08}.
%Type = Article
\bibitem[{Aron and Schwartz(1984)}]{aronSEIR}
\bibinfo{author}{J.~Aron}, \bibinfo{author}{I.~Schwartz},
\newblock \bibinfo{title}{Seasonality and period-doubling bifurcations in an
  epidemic model},
\newblock \bibinfo{journal}{Journal of theoretical biology}
  \bibinfo{volume}{110} (\bibinfo{year}{1984}) \bibinfo{pages}{665—679}.
  \DOIprefix\doi{10.1016/s0022-5193(84)80150-2}.
%Type = Article
\bibitem[{Kermack et~al.(1997)Kermack, McKendrick, and
  Walker}]{KermackMcKendrickWalker1997}
\bibinfo{author}{W.~O. Kermack}, \bibinfo{author}{A.~G. McKendrick},
  \bibinfo{author}{G.~T. Walker},
\newblock \bibinfo{title}{A contribution to the mathematical theory of
  epidemics},
\newblock \bibinfo{journal}{Proceedings of the Royal Society of London. Series
  A, Containing Papers of a Mathematical and Physical Character}
  \bibinfo{volume}{115} (\bibinfo{year}{1997}) \bibinfo{pages}{700–721}.
  \DOIprefix\doi{10.1098/rspa.1927.0118}.
%Type = Article
\bibitem[{Aparicio and Pascual(2006)}]{AparicioPascual2006}
\bibinfo{author}{J.~P. Aparicio}, \bibinfo{author}{M.~Pascual},
\newblock \bibinfo{title}{Building epidemiological models from r0: an implicit
  treatment of transmission in networks},
\newblock \bibinfo{journal}{Proceedings of the Royal Society B: Biological
  Sciences} \bibinfo{volume}{274} (\bibinfo{year}{2006})
  \bibinfo{pages}{505–512}. \DOIprefix\doi{10.1098/rspb.2006.0057}.
%Type = Article
\bibitem[{Hoen et~al.(2015)Hoen, Hladish, Eggo, Lenczner, Brownstein, and
  Meyers}]{HoenHladishEggoLencznerBrownsteinMeyers2015}
\bibinfo{author}{A.~G. Hoen}, \bibinfo{author}{T.~J. Hladish},
  \bibinfo{author}{R.~M. Eggo}, \bibinfo{author}{M.~Lenczner},
  \bibinfo{author}{J.~S. Brownstein}, \bibinfo{author}{L.~A. Meyers},
\newblock \bibinfo{title}{Epidemic wave dynamics attributable to urban
  community structure: A theoretical characterization of disease transmission
  in a large network},
\newblock \bibinfo{journal}{Journal of Medical Internet Research}
  \bibinfo{volume}{17} (\bibinfo{year}{2015}) \bibinfo{pages}{e169}.
  \DOIprefix\doi{10.2196/jmir.3720}.
%Type = Misc
\bibitem[{Xue(2020)}]{Xue2020}
\bibinfo{author}{S.~Xue}, \bibinfo{title}{Scheduling and online planning in
  stochastic diffusion networks}, \bibinfo{year}{2020}. \URLprefix
  \url{https://ir.library.oregonstate.edu/concern/graduate\_thesis\_or\_dissertations/k0698f81q}.
%Type = Inproceedings
\bibitem[{Kinathil et~al.(2017)Kinathil, Soh, and
  Sanner}]{kinathil2017nonlinear}
\bibinfo{author}{S.~Kinathil}, \bibinfo{author}{H.~Soh},
  \bibinfo{author}{S.~Sanner},
\newblock \bibinfo{title}{Nonlinear optimization and symbolic dynamic
  programming for parameterized hybrid markov decision processes},
\newblock in: \bibinfo{booktitle}{Workshops at the Thirty-First AAAI Conference
  on Artificial Intelligence}, \bibinfo{year}{2017}.
%Type = Article
\bibitem[{Arruda et~al.(2022)Arruda, Sharma, Alexandre, and
  Thomas}]{arruda2022}
\bibinfo{author}{E.~F. Arruda}, \bibinfo{author}{T.~Sharma},
  \bibinfo{author}{R.~e.~A. Alexandre}, \bibinfo{author}{S.~S. Thomas},
\newblock \bibinfo{title}{Epidemic control modeling using parsimonious models
  and markov decision processes}  (\bibinfo{year}{2022}). \URLprefix
  \url{http://arxiv.org/abs/2206.13910}.
  \DOIprefix\doi{10.48550/arXiv.2206.13910}, \bibinfo{note}{arXiv:2206.13910
  [physics, q-bio]}.
%Type = Misc
\bibitem[{Li et~al.(2021)Li, Haddadan, Li, Marathe, Srinivasan, Vullikanti, and
  Zhao}]{li2021mdpLP}
\bibinfo{author}{G.~Li}, \bibinfo{author}{A.~Haddadan},
  \bibinfo{author}{A.~Li}, \bibinfo{author}{M.~Marathe},
  \bibinfo{author}{A.~Srinivasan}, \bibinfo{author}{A.~Vullikanti},
  \bibinfo{author}{Z.~Zhao}, \bibinfo{title}{A markov decision process
  framework for efficient and implementable contact tracing and isolation},
  \bibinfo{year}{2021}. \URLprefix \url{https://arxiv.org/abs/2112.15547}.
  \DOIprefix\doi{10.48550/ARXIV.2112.15547}.
%Type = Inproceedings
\bibitem[{Lee et~al.(2021)Lee, Lee, Mogk, Goldstein, Bibliowicz, Brudy, and
  Tessier}]{LeeLeeMogkGoldsteinBibliowiczBrudyTessier2021}
\bibinfo{author}{B.~Lee}, \bibinfo{author}{M.~Lee}, \bibinfo{author}{J.~Mogk},
  \bibinfo{author}{R.~Goldstein}, \bibinfo{author}{J.~Bibliowicz},
  \bibinfo{author}{F.~Brudy}, \bibinfo{author}{A.~Tessier},
\newblock \bibinfo{title}{Designing a multi-agent occupant simulation system to
  support facility planning and analysis for covid-19},
\newblock in: \bibinfo{booktitle}{Designing Interactive Systems Conference
  2021}, DIS ’21, \bibinfo{publisher}{Association for Computing Machinery},
  \bibinfo{address}{New York, NY, USA}, \bibinfo{year}{2021}, p.
  \bibinfo{pages}{15–30}. \URLprefix
  \url{https://doi.org/10.1145/3461778.3462030}.
  \DOIprefix\doi{10.1145/3461778.3462030}.
%Type = Article
\bibitem[{Markhorst et~al.(2021)Markhorst, Zver, Malbasic, Dijkstra, Otto,
  van~der Mei, and Moeke}]{MarkhorstZverMalbasicDijkstraOttovanderMeiMoeke2021}
\bibinfo{author}{B.~Markhorst}, \bibinfo{author}{T.~Zver},
  \bibinfo{author}{N.~Malbasic}, \bibinfo{author}{R.~Dijkstra},
  \bibinfo{author}{D.~Otto}, \bibinfo{author}{R.~van~der Mei},
  \bibinfo{author}{D.~Moeke},
\newblock \bibinfo{title}{A data-driven digital application to enhance the
  capacity planning of the covid-19 vaccination process},
\newblock \bibinfo{journal}{Vaccines} \bibinfo{volume}{9}
  (\bibinfo{year}{2021}) \bibinfo{pages}{1181}.
  \DOIprefix\doi{10.3390/vaccines9101181}.
%Type = Inproceedings
\bibitem[{He et~al.(2022)He, Cao, and Ye}]{HeCaoYe2022}
\bibinfo{author}{W.~He}, \bibinfo{author}{Z.~Cao}, \bibinfo{author}{H.~Ye},
\newblock \bibinfo{title}{Path planning algorithms for mobile robots in
  hospital environment during covid-19},
\newblock in: \bibinfo{booktitle}{Proceedings of the 3rd International
  Symposium on Artificial Intelligence for Medicine Sciences}, ISAIMS ’22,
  \bibinfo{publisher}{Association for Computing Machinery},
  \bibinfo{address}{New York, NY, USA}, \bibinfo{year}{2022}, p.
  \bibinfo{pages}{522–530}. \URLprefix
  \url{https://doi.org/10.1145/3570773.3570853}.
  \DOIprefix\doi{10.1145/3570773.3570853}.
%Type = Article
\bibitem[{Barnawi et~al.(2021)Barnawi, Chhikara, Tekchandani, Kumar, and
  Boulares}]{BarnawiChhikaraTekchandaniKumarBoulares2021}
\bibinfo{author}{A.~Barnawi}, \bibinfo{author}{P.~Chhikara},
  \bibinfo{author}{R.~Tekchandani}, \bibinfo{author}{N.~Kumar},
  \bibinfo{author}{M.~Boulares},
\newblock \bibinfo{title}{A cnn-based scheme for covid-19 detection with
  emergency services provisions using an optimal path planning},
\newblock \bibinfo{journal}{Multimedia Systems}  (\bibinfo{year}{2021}).
  \URLprefix \url{https://doi.org/10.1007/s00530-021-00833-2}.
  \DOIprefix\doi{10.1007/s00530-021-00833-2}.
%Type = Unpublished
\bibitem[{Sanner(2010)}]{Sanner:RDDL}
\bibinfo{author}{S.~Sanner}, \bibinfo{title}{Relational dynamic influence
  diagram language (rddl): Language description}, \bibinfo{year}{2010}.
%Type = Misc
\bibitem[{CDC(????)}]{cdc}
\bibinfo{author}{CDC}, \bibinfo{title}{Isolation and precautions for people
  with covid-19}, ???? \URLprefix
  \url{https://www.cdc.gov/coronavirus/2019-ncov/your-health/isolation.html}.
%Type = Article
\bibitem[{Haischer et~al.(2020)Haischer, Beilfuss, Hart, Opielinski, Wrucke,
  Zirgaitis, Uhrich, and Hunter}]{haischer2020wearing}
\bibinfo{author}{M.~H. Haischer}, \bibinfo{author}{R.~Beilfuss},
  \bibinfo{author}{M.~R. Hart}, \bibinfo{author}{L.~Opielinski},
  \bibinfo{author}{D.~Wrucke}, \bibinfo{author}{G.~Zirgaitis},
  \bibinfo{author}{T.~D. Uhrich}, \bibinfo{author}{S.~K. Hunter},
\newblock \bibinfo{title}{Who is wearing a mask? gender-, age-, and
  location-related differences during the covid-19 pandemic},
\newblock \bibinfo{journal}{PloS one} \bibinfo{volume}{15}
  (\bibinfo{year}{2020}) \bibinfo{pages}{e0240785}.
%Type = Article
\bibitem[{Gravelle et~al.(2022)Gravelle, Phillips, Reifler, and
  Scotto}]{gravelle2022estimating}
\bibinfo{author}{T.~B. Gravelle}, \bibinfo{author}{J.~B. Phillips},
  \bibinfo{author}{J.~Reifler}, \bibinfo{author}{T.~J. Scotto},
\newblock \bibinfo{title}{Estimating the size of “anti-vax” and vaccine
  hesitant populations in the us, uk, and canada: comparative latent class
  modeling of vaccine attitudes},
\newblock \bibinfo{journal}{Human vaccines \& immunotherapeutics}
  \bibinfo{volume}{18} (\bibinfo{year}{2022}) \bibinfo{pages}{2008214}.
%Type = Article
\bibitem[{Lewis et~al.(2022)Lewis, Chambers, Chu, Fortnam, De~Vito, Gargano,
  Chan, McDonald, and Hogan}]{lewis2022effectiveness}
\bibinfo{author}{N.~Lewis}, \bibinfo{author}{L.~C. Chambers},
  \bibinfo{author}{H.~T. Chu}, \bibinfo{author}{T.~Fortnam},
  \bibinfo{author}{R.~De~Vito}, \bibinfo{author}{L.~M. Gargano},
  \bibinfo{author}{P.~A. Chan}, \bibinfo{author}{J.~McDonald},
  \bibinfo{author}{J.~W. Hogan},
\newblock \bibinfo{title}{Effectiveness associated with vaccination after
  covid-19 recovery in preventing reinfection},
\newblock \bibinfo{journal}{JAMA network open} \bibinfo{volume}{5}
  (\bibinfo{year}{2022}) \bibinfo{pages}{e2223917--e2223917}.
%Type = Misc
\bibitem[{of~Ontario(2020)}]{ontarioschools}
\bibinfo{author}{G.~of~Ontario}, \bibinfo{title}{Schools covid-19 data -
  ontario data catalogue}, \bibinfo{year}{2020}. \URLprefix
  \url{https://data.ontario.ca/en/dataset/summary-of-cases-in-schools}.
%Type = Misc
\bibitem[{for Health~Information(2023)}]{cihisite}
\bibinfo{author}{C.~I. for Health~Information}, \bibinfo{title}{Canadian
  covid-19 intervention timeline}, \bibinfo{year}{2023}. \URLprefix
  \url{https://www.cihi.ca/en/canadian-covid-19-intervention-timeline}.
%Type = Misc
\bibitem[{for Health~Information(2022)}]{cihi}
\bibinfo{author}{C.~I. for Health~Information}, \bibinfo{title}{Canadian data
  set of covid-19 interventions — data tables}, \bibinfo{year}{2022}.
  \URLprefix
  \url{https://www.cihi.ca/sites/default/files/document/covid-19-intervention-scan-data-tables-en.xlsx}.
%Type = Misc
\bibitem[{of~Ontario(2023)}]{ontarioeducationfinder}
\bibinfo{author}{G.~of~Ontario}, \bibinfo{title}{School information finder},
  \bibinfo{year}{2023}. \URLprefix
  \url{https://www.app.edu.gov.on.ca/eng/sift/index.asp}.

\end{thebibliography}
\end{document}